\documentclass[runningheads]{llncs}
\usepackage[T1]{fontenc}
\usepackage{graphicx}
\usepackage{subcaption}
\usepackage{amsmath}
\usepackage[linesnumbered,ruled,vlined]{algorithm2e}
\usepackage{multirow}
\usepackage{threeparttable} 
\usepackage[normalem]{ulem}

\usepackage[aboveskip=5pt, belowskip=5pt]{caption}       
\usepackage{caption}
\usepackage{subcaption}
\usepackage{xcolor}
\begin{document}
%
\title{HOTA: Hierarchical Overlap-Tiling Aggregation for Large-Area 3D Flood Mapping}

\titlerunning{HOTA for 3D Flood Mapping}
%
\author{Wenfeng Jia\inst{1}\orcidID{0000-0002-3996-5438}  \and
Bin Liang\inst{2}\orcidID{0000-0002-6605-2167} \and
Yuxi Lu\inst{2}\orcidID{0009-0002-2220-7873} \and
Attavit Wilaiwongsakul\inst{2}\orcidID{0009-0003-3084-9419} \and
Muhammad Arif Khan\inst{1}\orcidID{0000-0001-6112-8874} \and
Lihong Zheng\thanks{Corresponding author: lzheng@csu.edu.au}\inst{1}\orcidID{0000-0001-5728-4356}}
\authorrunning{W. Jia et al.}
%
\institute{Charles Sturt University, New South Wales, Australia  \and
University of Technology Sydney, New South Wales, Australia}
\maketitle              
\begin{abstract}
Floods are among the most frequent natural hazards and cause significant social and economic damage. Timely, large-scale information on flood extent and depth is essential for disaster response; however, existing products often trade spatial detail for coverage or ignore flood depth altogether. 
To bridge this gap, this work presents HOTA: Hierarchical Overlap-Tiling Aggregation, a plug-and-play, multi-scale inference strategy. 
When combined with SegFormer and a dual-constraint depth estimation module, this approach forms a complete 3D flood-mapping pipeline.
HOTA applies overlapping tiles of different sizes to multispectral Sentinel-2 images only during inference, enabling the SegFormer model to capture both local features and kilometre-scale inundation without changing the network weights or retraining. 
The subsequent depth module is based on a digital elevation model (DEM) differencing method,  which refines the 2D mask and estimates flood depth by enforcing (i) zero depth along the flood boundary and (ii) near-constant flood volume with respect to the DEM.
A case study on the March 2021 Kempsey (Australia) flood shows that HOTA, when coupled with SegFormer, improves IoU from 73\% (U-Net baseline) to 84\%. The resulting 3D surface achieves a mean absolute boundary error of less than 0.5 m. 
These results demonstrate that HOTA can produce accurate, large-area 3D flood maps suitable for rapid disaster response.
\keywords{Flood Mapping  \and Deep Learning \and Remote Sensing \and 3D.}
\end{abstract}
%
%
\section{Introduction}
\begin{figure}[htbp]
    \centering
    \includegraphics[width=0.65\textwidth]{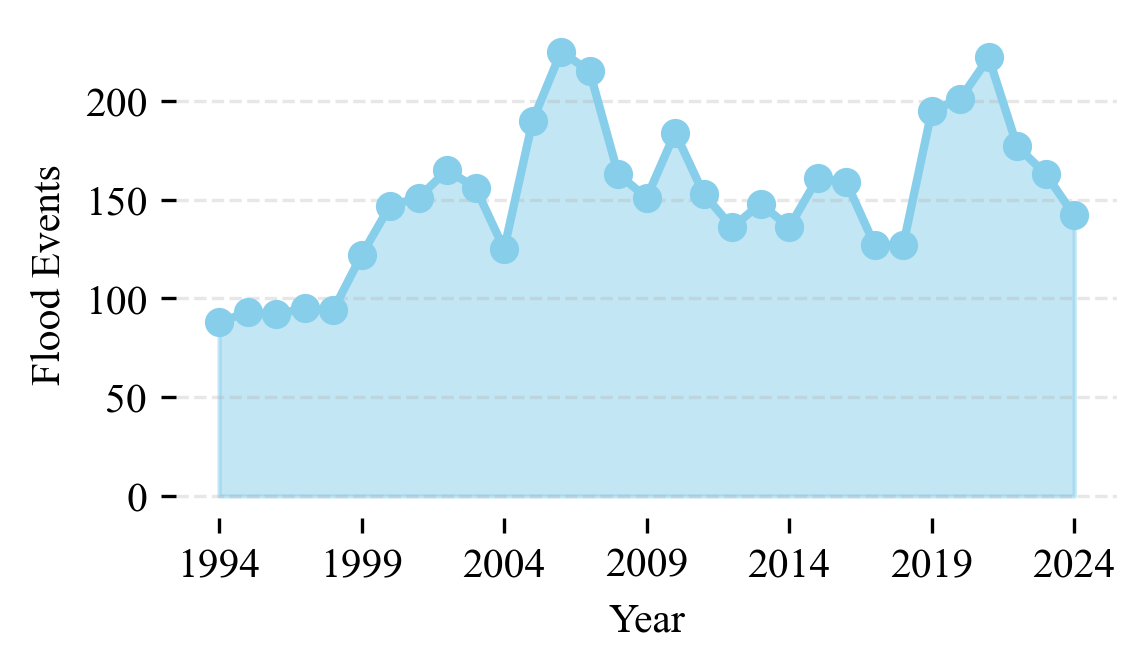}
    \caption{Flood events from 1994 to 2024.}
    \label{fig:floodEvents}
\end{figure}

Flooding has long been one of the most devastating natural disasters, posing serious threats to both human life and economic stability~\cite{Flood2022affect,Flood2022increase}. 
According to statistics from the Emergency Events Database (EM-DAT)~\cite{EM-DAT2025}, as shown in Fig.~\ref{fig:floodEvents}, the average annual number of recorded flood events in the 2020s reached over 150, nearly doubling the average of 75 events per year in the 1990s. 
Between 2018 and 2022 alone, a total of 807 flood events were reported worldwide, resulting in over 20,000 deaths and disappearances, as well as more than USD 180 billion in direct economic losses~\cite{Li2024Flood_DL_review}. 
To safeguard human lives and promote global economic resilience, stakeholders and policymakers must develop effective flood mitigation and management strategies.

Flood mapping is a fundamental and widely used approach for flood prevention and mitigation~\cite{Sun2025FloodMapping}. Traditionally, it has relied on two-dimensional (2D) representations to delineate inundated areas on a flat plane. 
However, with the accelerating pace of urbanization, the misuse of land and growing vulnerability of urban infrastructure have contributed to increasingly complex spatial patterns of flood risk~\cite{Wedajo2024TwoD_drawbacks}. 
As a result, conventional 2D flood mapping tools are no longer sufficient for addressing the spatial challenges posed by modern flood events~\cite{Percival2020TwoD_drawbacks}. 
To overcome the limitations of traditional flood mapping, three-dimensional (3D) flood mapping has emerged as a promising alternative~\cite{Jia2025Survey3DFlood,Tan2024upscalingDEM_CNN}. 
By integrating high-resolution terrain data, hydrodynamic models and computer graphics technologies, 3D flood mapping can provide detailed visualizations of flood extent, depth, and dynamic changes over time.

In this study, we proposes \textbf{HOTA}, Hierarchical Overlap-Tiling Aggregation, a plug-and-play multi-scale inference strategy. And bulid a 3D flood mapping pipeline by combining HOTA with a dual-constraint Digital Elevation Model (DEM) differencing module. 
First, we perform 2D flood mapping using the SegFormer model in combination with HOTA strategy, and compare its performance with the conventional U-Net model. 
Based on the 2D mapping results, we further introduce a dual-constraint differential water depth estimation method to enhance the accuracy of flood boundary and depth estimation. 
Finally, the effectiveness of the proposed method is validated using the March 2021 flood event in Kempsey, Australia.
The key contributions of this work are as follows:
\begin{enumerate}
    \item A comparative evaluation of SegFormer and U-Net for 2D flood mapping.
    \item The \textbf{HOTA} strategy, which is suitable for large-scale remote sensing tasks.
    \item A dual-constraint differential water depth estimation method to improve the accuracy of flood boundary and depth estimation.
\end{enumerate}

\section{Related Work}
Flood mapping has long relied on physically simulated hydrodynamic models~\cite{Karim2023HydrodynamicML_review}. 
Traditional methods typically use 2D models to simulate the flood evolution process. 
Representative tools include HEC-RAS, LISFLOOD-FP~\cite{Shaw2021LISFLOOD_FP}, and others. 
These models integrate terrain, roughness, and boundary conditions to produce 2D inundation maps or water depth contours with relatively high computational efficiency. 
However, by neglecting vertical dynamics, 2D models fail to accurately represent complex flood behavior. 
Although 3D hydrodynamic models provide more detailed flow simulations, they require extensive computational resources and are difficult to apply efficiently over large areas.

To improve efficiency, studies have proposed hybrid modeling approaches that combine machine learning with hydrodynamic models~\cite{Karim2023HydrodynamicML_review}. 
In recent years, the rapid growth of remote sensing data and computing power has further driven the development of deep learning-based flood mapping methods, which have become a new trend in 3D flood modeling~\cite{han2023survey,Jia2025Survey3DFlood,Seleem2023CNN_RF_Compare}.

\subsection{Research States for Flood Mapping}
Within deep learning methods, researchers have proposed 2 main strategies~\cite{Jia2025Survey3DFlood}.
First one is \textbf{task-decomposed} approach, which divides the 3D flood modeling process into data preprocessing, 2D prediction, and 3D depth estimation. 
These tasks are handled by deep learning models or physical models respectively, and the outputs are fused to generate the final 3D flood map. 
This approach offers good interpretability and modular flexibility, but the process is complex and prone to error propagation between modules.

\textbf{End-to-End} is another approach, which trains a unified model to map input data, such as rainfall, DEMs, or remote sensing images, directly to flood depth maps. 
This approach enables direct depth prediction and is suitable for real-time applications. 
However, its generalization ability is limited because this approach mainly relies on simulated datasets.
If the hydrodynamic model does not provide sufficient datasets, the model may produce inaccurate results in unseen areas. 

In summary, flood mapping techniques are gradually shifting from physics-driven to data-driven paradigms, and deep learning methods are becoming a key direction for 3D flood mapping.

\subsection{Deep Learning Models\& Flood Depth Estimation Methods}
In recent years, convolutional neural networks (CNNs) have been widely used in flood mapping, extracting spatial features from remote sensing imagery to segment flooded areas~\cite{Bentivoglio2022DL_model_review}. 
U-Net remains a mainstream model due to its multi-scale fusion and ability to preserve boundary details.
Some works further enhance feature representation with attention mechanisms and residual modules; for example, CRU-Net has shown strong performance in urban flood modeling~\cite{Shao2024CRU}.
For temporal modeling, Long Short-Term Memory (LSTM) networks are commonly used to capture dynamic relationships among rainfall, runoff, and water level. 
ConvLSTM combines convolutional and recurrent structures to model spatiotemporal flood evolution effectively~\cite{Dehghani2023ConvLSTM}.
Besides CNNs and LSTM, deep learning models like GNNs~\cite{Bentivoglio2023SWE-GNN}, GANs~\cite{do2023CGAN} also demonstrate the giant potential of deep learning on flood mapping.

\begin{figure}[htbp]
    \centering
    \includegraphics[width=0.45\textwidth]{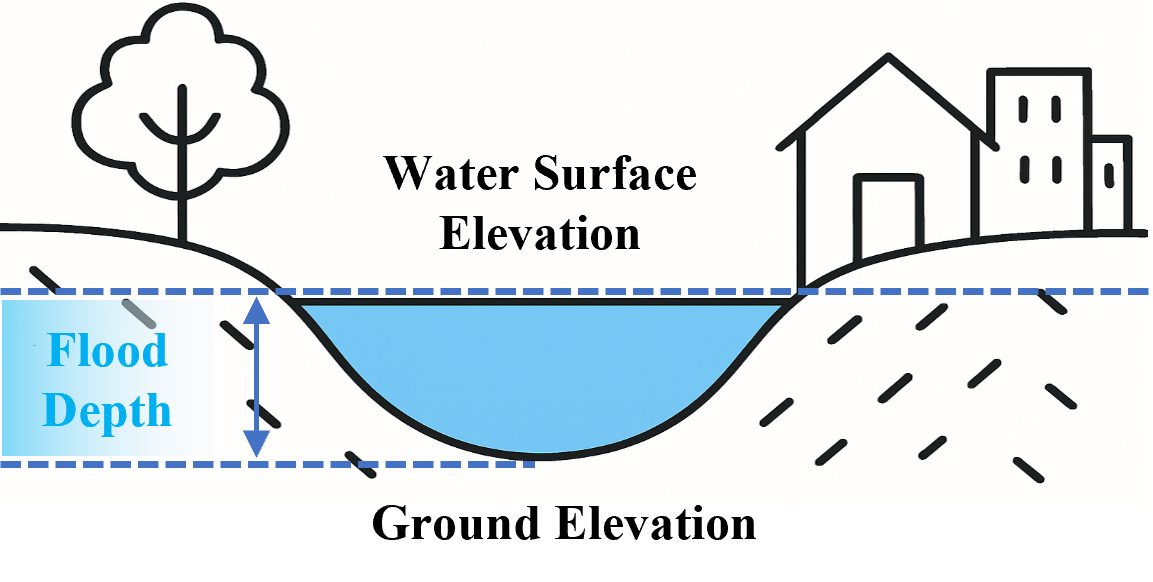}
    \caption{Difference method for flood depth estimation.}
    \label{fig:Difference}
\end{figure}

While flood depth estimation is essential for 3D flood modeling, most of deep learning models in task-decomposed approaches are used to predict 2D flooded areas~\cite{Jia2025Survey3DFlood}.
Flood depth prediction is primarily performed using the differential method~\cite{Betterle2024FloodDepth,Poterek2025INFLOS,Tan2024upscalingDEM_CNN} (shown in Eq.~\ref{eq:demDifference} and Fig.~\ref{fig:Difference}).
The differential method extracts flood boundaries elevations, estimates water surface elevation, and subtracts DEM-based ground elevation to compute flood depth. 
It is fast and suitable for simple terrain but sensitive to boundary and DEM accuracy. 
\begin{equation}
    \text{Flood Depth} = \text{Water Surface Elevation} - \text{Ground Elevation}
\label{eq:demDifference}
\end{equation}

Despite developments in 2D and 3D flood mapping, challenges remain. 
Existing methods often miss multi-scale flood features, failing to capture both narrow boundaries and large-scale flood bodies. 
Also, standard vision methods overlook the large-scale nature of floods in remote sensing, causing tile-edge discontinuities. 
Moreover, depth estimation using the differential method is limited by the accuracy of 2D boundaries, and errors in these boundaries propagate to depth predictions.
To address these issues, we propose a 3D flood mapping pipeline based on \textbf{HOTA}. 
The technical workflow and key components are detailed in the next section.

\section{Methodology}
\label{sec:Methodology}
\begin{figure}[htbp]
    \centering
    \includegraphics[width=0.85\textwidth]{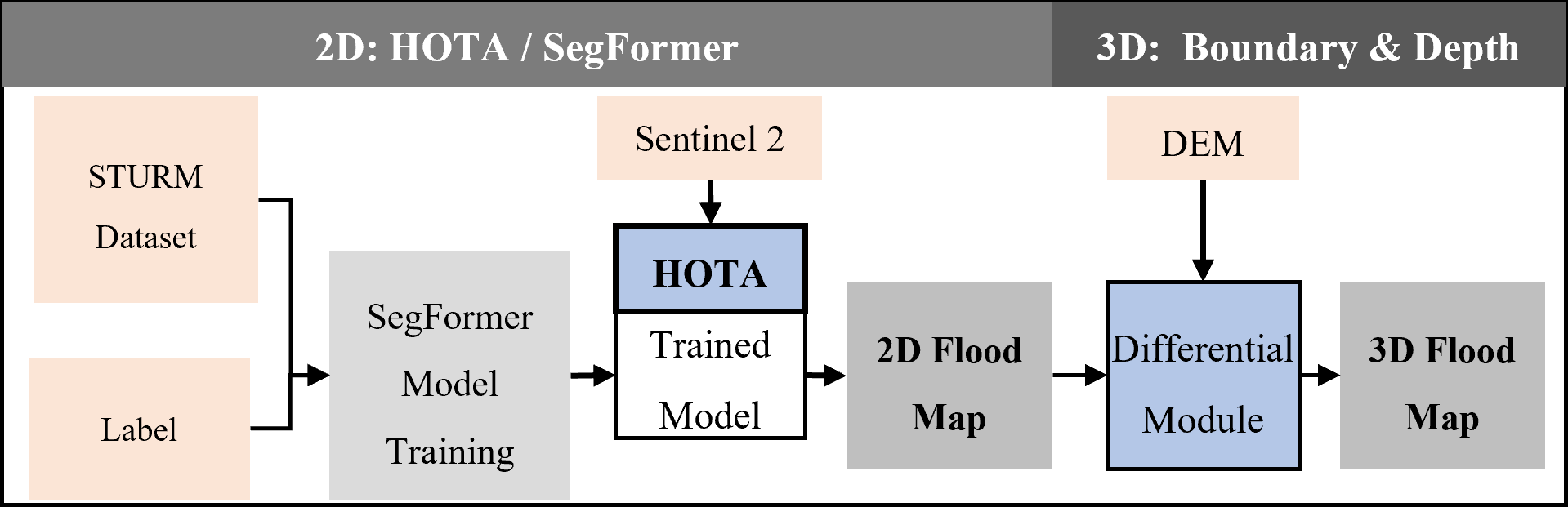}
    \caption{Complete technical workflow.}
    \label{fig:Methodology}
\end{figure}

This section first provides an overview of the complete technical workflow (see Fig.~\ref{fig:Methodology}), then describes each key component in detail, 
including the study area and datasets (Sec.~\ref{sec:area}), 
deep learning model training (Sec.~\ref{sec:dltrain}), 
the Hierarchical Overlap-Tiling Aggregation (\textbf{HOTA}) inference strategy (Sec.~\ref{sec:hota}), 
and the DEM-based flood depth estimation and boundary refinement (Sec.\ref{sec:depth}).

The overall workflow of this study adopts a task decomposition approach, dividing the process into 2D and 3D components. 
The 2D module focuses on flood segmentation using SegFormer and HOTA strategy, and includes: 
(i) definition of the study area and dataset construction, 
(ii) building and training the SegFormer model, 
and (iii) implementation of the HOTA inference strategy. 
The 3D component involves flood depth estimation and boundary refinement based on the integration of a Digital Elevation Model (DEM) and the 2D flood maps. 
All modules operate in sequence to ultimately generate a 3D flood map.

\subsection{Research Area \& Datasets}
\label{sec:area}
\begin{figure}[htbp]
    \centering
    \begin{subfigure}[b]{0.32\textwidth}
        \centering
        \includegraphics[width=\textwidth]{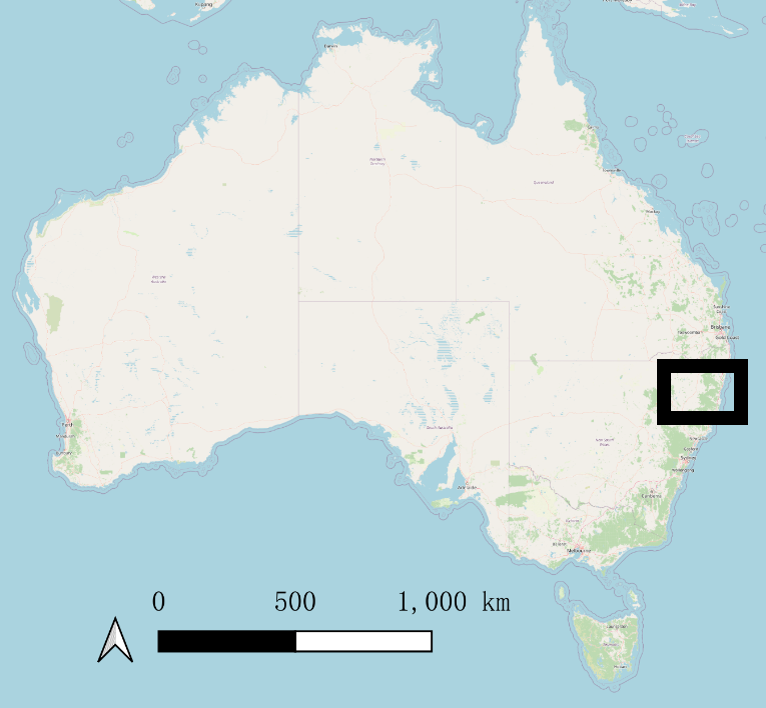}
        \caption{Study area location}
        \label{fig:kempsey_location}
    \end{subfigure}%
    \hspace{0.5em} %
    \begin{subfigure}[b]{0.32\textwidth}
        \centering
        \includegraphics[width=\textwidth]{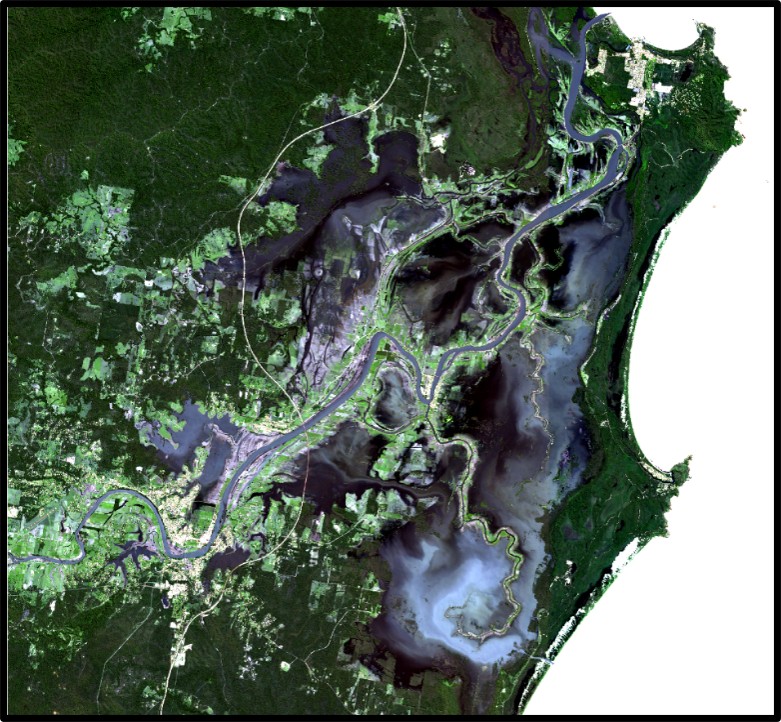}
        \caption{Sentinel-2 RGB image}
        \label{fig:kempsey_s2}
    \end{subfigure}
    \begin{subfigure}[b]{0.38\textwidth}
        \centering
        \includegraphics[width=\textwidth]{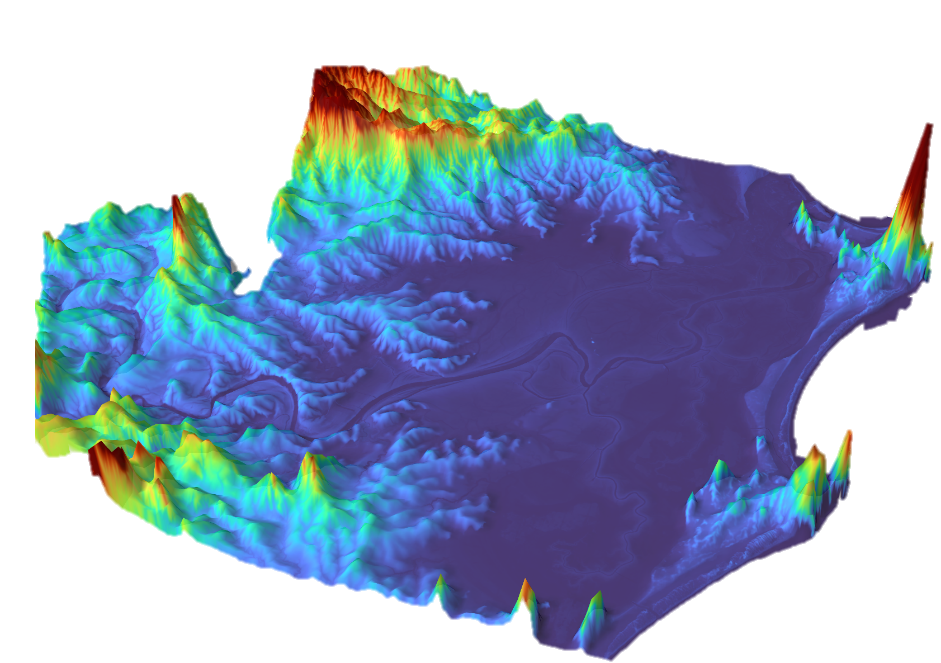}
        \caption{DEM}
        \label{fig:kempsey_dem}
    \end{subfigure}
    \begin{subfigure}[b]{0.31\textwidth}
        \centering
        \includegraphics[width=\textwidth]{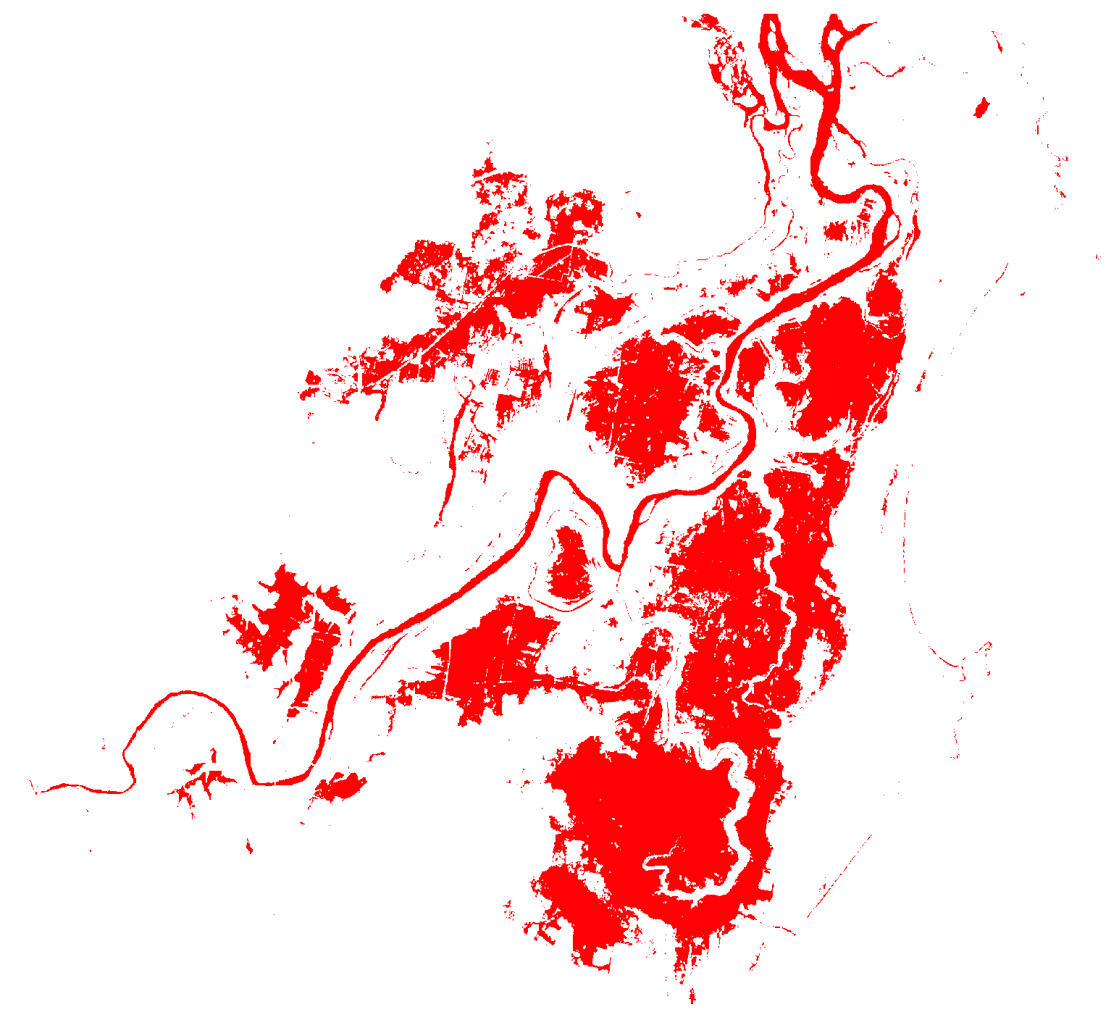}
        \caption{2D ground truth}
        \label{fig:kempsey_gt}
    \end{subfigure}
    \caption{Overview of the Kempsey flood dataset.}
    \label{fig:kempsey}
\end{figure}

The dataset in this study consists of two main parts: 
(i) the STURM-Flood dataset (Sentinel-2, S2)~\cite{Notarangelo2025STURM}, which is used to train the SegFormer and U-Net (baseline) models, 
and (ii) a case study of a flood event in Kempsey, New South Wales, Australia. The Kempsey dataset (Fig.~\ref{fig:kempsey}) serves as a benchmark for model comparison, evaluation of the HOTA strategy, and subsequent 3D processing.

The STURM-Flood (S2) comprises 2,675 Sentinel-2 tiles of (H$\times$W) 128$\times$128 pixels at 10 m spatial resolution, each paired with a binary water mask covering 60 flood events worldwide.
It provides high-quality, atmospherically corrected multispectral imagery specifically prepared for training and evaluating deep learning models in optical flood extent mapping using Sentinel-2 data~\cite{Notarangelo2025STURM}.

The case study focuses on the flood event that occurred in Kempsey, Australia (see Fig.~\ref{fig:kempsey_location}) in March 2021, triggered by heavy rainfall. 
The Kempsey dataset used in this study includes S2 satellite imagery, a 2D flood map ground truth, and a high-resolution DEM.

Due to the constraints of satellite revisit cycles and cloud cover, S2 imagery acquired on 26 March 2021 was selected for analysis. 
The flood image contains 9 spectral bands, consistent with the STURM dataset (S2). 
Among these, 4 bands (B2, B3, B4, B8) have a spatial resolution of 10m, while the remaining 5 bands (B5, B6, B7, B11, B12) were originally at 20m resolution and subsequently resampled to 10m. 
The image size exceeds 3900$\times$3600 pixels. Fig.~\ref{fig:kempsey_s2} shows the RGB composite of the flood situation.
Additionally, the Sentinel-2 Scene Classification Layer (SCL) was used for water detection.
In this study, the water class (SCL=6) was extracted from the SCL map to generate the 2D ground truth (see Fig.~\ref{fig:kempsey_gt}). 
Since the SCL product is provided at 20m resolution, it was also resampled to 10m to align with the other data.

The DEM (Fig.~\ref{fig:kempsey_dem}) used for flood depth estimation was obtained from the 5m resolution LiDAR-derived DEM product released by Geoscience Australia. 
This product meets the Australian ICSM LiDAR Acquisition Specifications, providing a vertical accuracy better than 0.3m (95\% confidence). 
No further preprocessing was required, as the DEM reliably represents the terrain characteristics of the study area and provides a robust foundation for flood depth estimation.

\subsection{SegFormer and U-Net Building \& Training}
\label{sec:dltrain}
\begin{figure}[htbp]
    \centering
    \includegraphics[width=\textwidth]{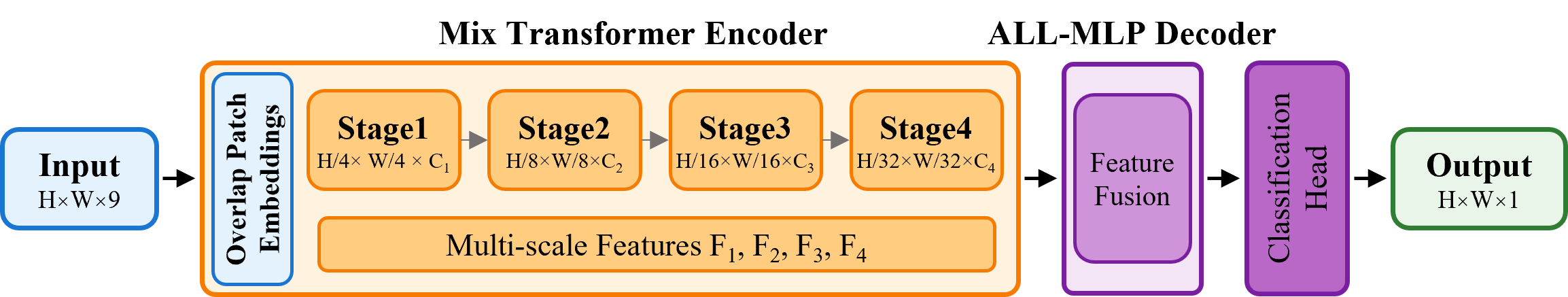}
    \caption{SegFormer model.}
    \label{fig:segformer}
\end{figure}

In this study, 2D segmentation is performed using the SegFormer model (shown in Fig.~\ref{fig:segformer}), with the U-Net serving as the baseline. 
SegFormer~\cite{Xie2021segformer} employs a MiT (Mix Transformer) architecture  that integrates the strengths of both Transformer and CNNs, providing strong global feature learning capabilities.

This model uses MiT-B2 as the encoder, and ImageNet pre-trained weights are applied. 
The input consists of 9 bands from the S2 data, and the output is a binary classification of water versus non-water. 
To ensure efficient and stable model training, the loss function is a weighted combination of binary cross-entropy (BCE) and Dice coefficient, with weights of 0.7 and 0.3, respectively. 
And the AdamW optimizer is used, with a learning rate set to $1 \times 10^{-4}$, a batch size of 64, and 200 training epochs.

For comparison, the baseline U-Net adopts the same training parameters, but uses the CNN-based ResNet34 as the encoder (the same parameter scale of SegFormer), facilitating comparison with previous studies~\cite{Notarangelo2025STURM}. 
Both models are trained on the STURM S2 dataset, with the dataset split into training, validation, and test sets in a ratio of 0.8 : 0.1 : 0.1.

\subsection{Hierarchical Overlap-Tiling Aggregation}
\label{sec:hota}

Despite the demonstrated global feature modeling capability of SegFormer in other domains, two main challenges remain in its application to flood mapping. 
First, due to the large spatial extent of floods, water boundaries often intersect with the edges of tiles after the image is divided, causing discontinuities or omissions along flood boundaries. 
Second, a single-scale sliding window cannot effectively capture both narrow channels and extensive floodplains.

\begin{figure}[htbp]
    \centering
    \begin{subfigure}[b]{0.29\textwidth}
        \centering
        \includegraphics[width=\textwidth]{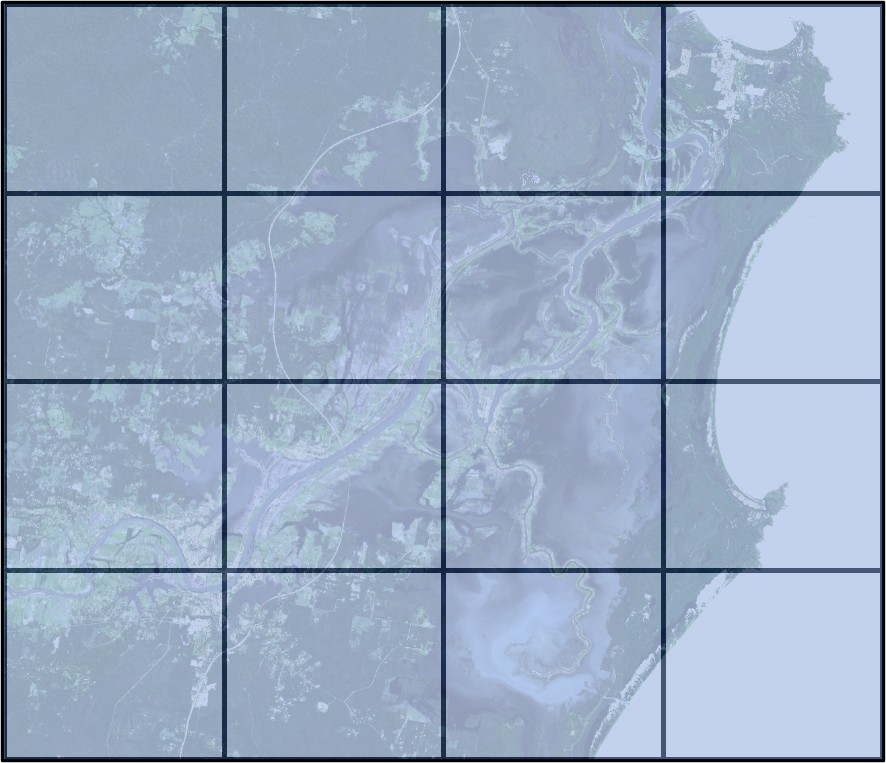}
        \caption{Conventional tiling}
        \label{fig:Tiling_ct}
    \end{subfigure}%
    \hspace{0em}
    \begin{subfigure}[b]{0.35\textwidth}
        \centering
        \includegraphics[width=\textwidth]{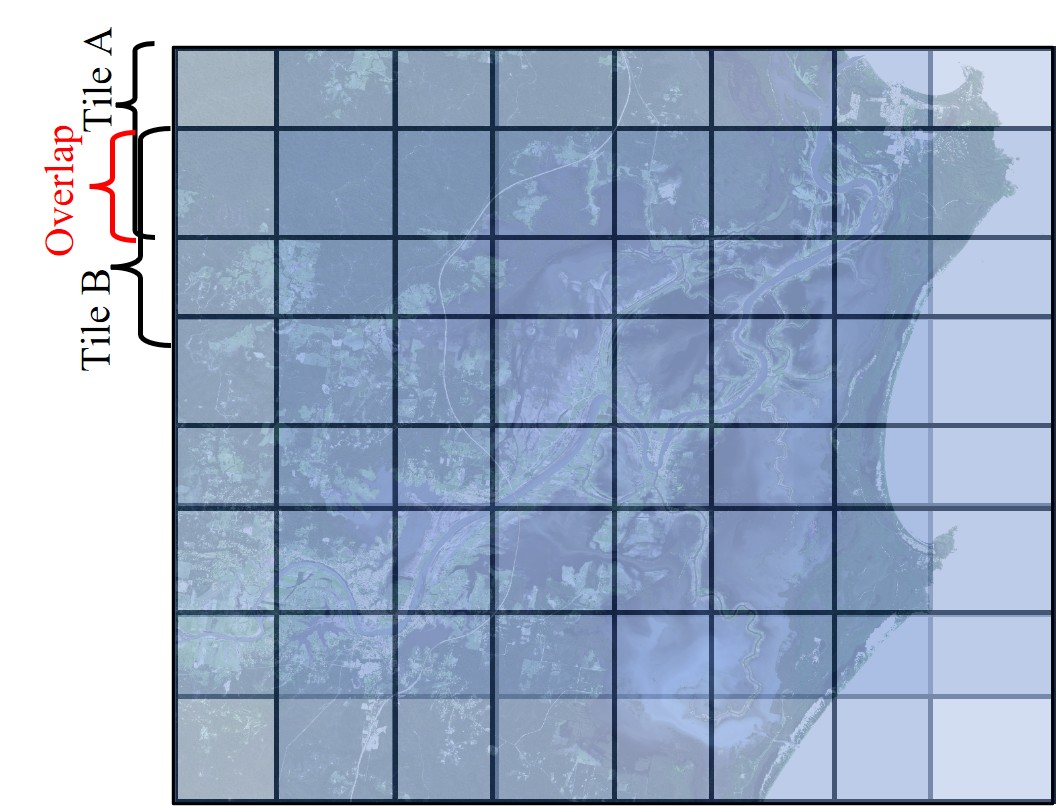}
        \caption{Overlap-tiling}
        \label{fig:Tiling_ot}
    \end{subfigure}
    \begin{subfigure}[b]{0.3\textwidth}
        \centering
        \includegraphics[width=\textwidth]{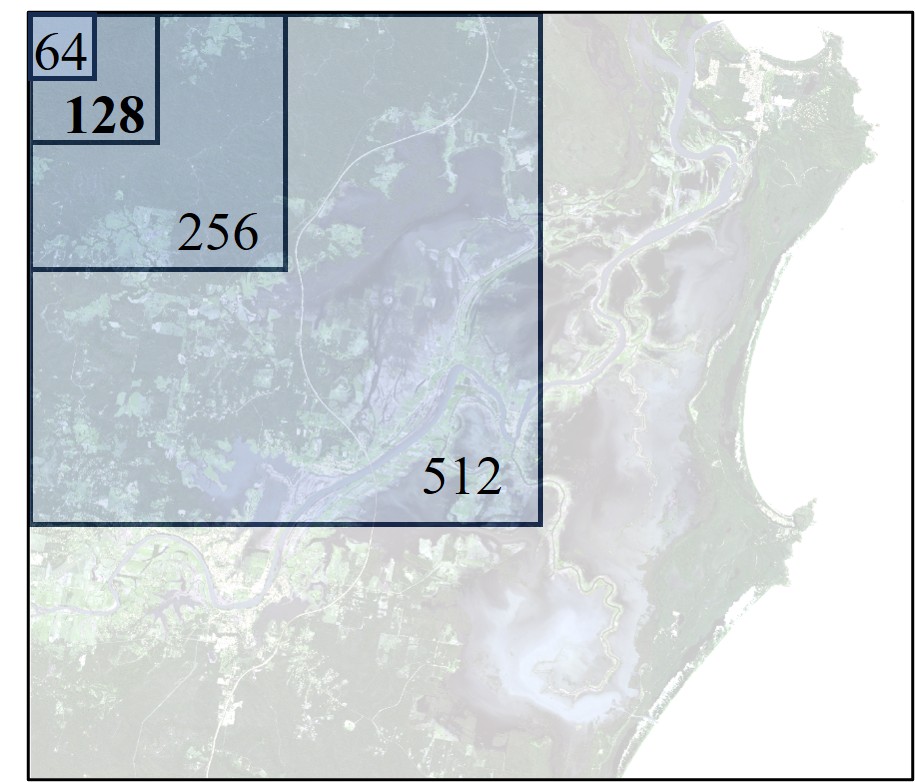}
        \caption{Hierarchical tiling}
        \label{fig:Tiling_ht}
    \end{subfigure}
    \caption{Hierarchical Overlap-Tiling.}
    \label{fig:Tiling}
\end{figure}

To address these issues, this study introduces a Hierarchical Overlap-Tiling Aggregation (HOTA) strategy during the inference stage. 
HOTA applies a multi-scale sliding window approach with window sizes of 64, 128, 256, and 512 pixels, combined with overlapping windows (e.g., 50\% overlap). 
This design ensures that each pixel receives abundant and diverse global context information by appearing in multiple tiles of different sizes and positions.
The core concept is illustrated in Fig.~\ref{fig:Tiling}, where Fig.~\ref{fig:Tiling_ot} shows the tiling configuration with 50\% overlap, and Fig.~\ref{fig:Tiling_ht} demonstrates the effects of different tile scales.
During the tile prediction fusion phase, a confidence-based fusion method is used to integrate overlapping predictions, which improves boundary continuity and segmentation robustness. 

\begin{figure}[htbp]
    \centering
    \begin{subfigure}[b]{0.25\textwidth}
        \centering
        \includegraphics[width=0.8\textwidth]{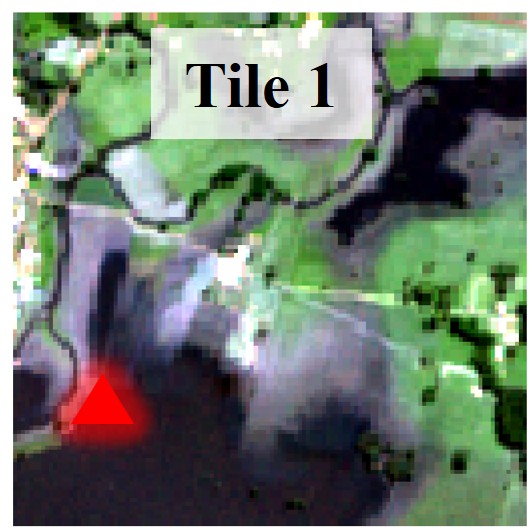}
        \caption{Conventional tiles}
        \label{fig:hotaTiling_ct}
    \end{subfigure}%
    \hspace{0em}
    \begin{subfigure}[b]{0.3\textwidth}
        \centering
        \includegraphics[width=0.8\textwidth]{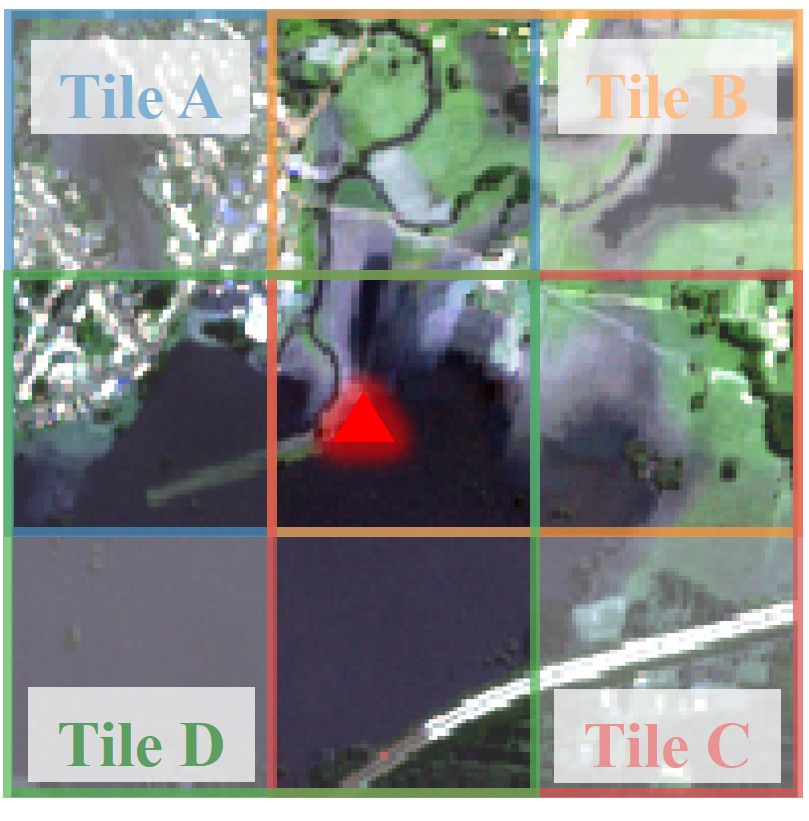}
        \caption{Overlapping tiles}
        \label{fig:hotaTiling_ot}
    \end{subfigure}
    \begin{subfigure}[b]{0.35\textwidth}
        \centering
        \includegraphics[width=0.9\textwidth]{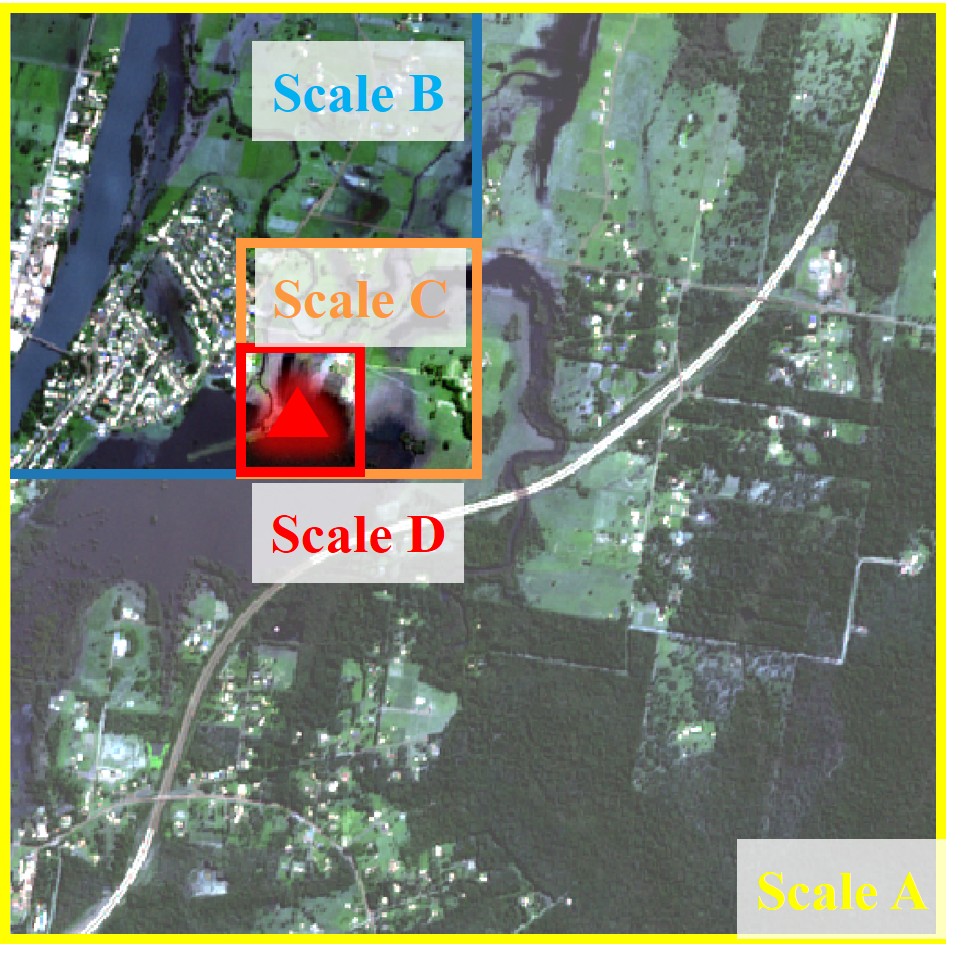}
        \caption{Hierarchical tiles}
        \label{fig:hotaTiling_ht}
    \end{subfigure}
    \caption{HOTA example.}
    \label{fig:hotaTiling}
\end{figure}

Fig.~\ref{fig:hotaTiling} presents an example of the HOTA strategy. It demonstrates how HOTA enables each location to receive diverse global information from up to different tiles. 
This approach allows for more accurate segmentation of floods, which often present multi-scale and complex structures. Specifically, Fig.~\ref{fig:hotaTiling_ct} displays the result of conventional tiling, Fig.~\ref{fig:hotaTiling_ot} shows the result with 50\% overlap, and Fig.~\ref{fig:hotaTiling_ht} depicts multi-scale tiling.

Compared to U-Net, SegFormer's MiT architecture is naturally robust to varying input sizes and well-suited for multi-scale inference. 
The combination of SegFormer and HOTA yields more accurate flood segmentation masks, providing a solid foundation for subsequent 3D flood depth modeling.

\subsection{Dual-Constraint Flood Depth and Boundary Joint Optimization}
\label{sec:depth}

While HOTA in the previous section enables high-quality 2D flood masks, accurate disaster assessment and risk management require 3D flood depth information.
To address this need, this subsection introduces a joint optimization (found in Algo.~\ref{alg:joint_depth_boundary}) that integrates DEM elevation and slope data with dual physical constraints to refine both flood depth and boundaries.
\begin{algorithm}[htbp]
    \caption{Flood Depth-Boundary Joint Optimization}
    \label{alg:joint_depth_boundary}
    \KwIn{Mask $M$, DEM $E$, volume tol. $\epsilon$, depth tol. $\delta$, max iter $N$}
    \KwOut{Depth $D^*$, Mask $M^*$}
    Morph. smooth $M \rightarrow M^*$; \\
    Estimate $W$ from DEM at $M^*$ boundary; \\
    $D \leftarrow \max(0, W - E)$ in $M^*$; \\
    $V_0 \leftarrow$ sum of $D$ in $M^*$; \\
    
    \For{$n = 1$ \KwTo $N$}{
        Estimate $W$ from DEM at $M^*$ boundary; \\
        $D \leftarrow \max(0, W - E)$ in $M^*$; \\
        $V \leftarrow$ sum of $D$ in $M^*$; \\
        \If{mean $|D|$ at boundary $< \delta$}{ break }
        \eIf{Boundary depth $>0$ or DEM slope outward}{
            Expand $M^*$;
        }{
            Contract $M^*$;
        }
        \If{$|V - V_0| > \epsilon V_0$}{ break }
        $V_0 \leftarrow V$;
    }
    \Return $D^*, M^*$
\end{algorithm}

The process begins by estimating initial depth through DEM and 2D map differencing (see Eq.~\ref{eq:demDifference}). 
Recognizing that the initial result may contain errors from segmentation, occlusions, or complex terrain, the flood boundary and depth are then jointly optimized. 
In each iteration, DEM elevation and slope guide the selective expansion or contraction of the flood extent, allowing inundation to spread into adjacent low-lying areas while excluding high-elevation regions.

Two physical constraints are imposed throughout the optimization: 
(i) water depth at the flood boundary is enforced to approach zero; 
and (ii) the flood volume must remain within a small, predefined range to prevent unrealistic expansion or contraction of the water body. 
The joint update of flood extent and water surface depth continues until the constraints are satisfied or a maximum number of iterations is reached. 
This produces a smooth, physically plausible 3D flood surface with improved boundary accuracy.

\section{Results and Analysis}
Based on the methodology described above, this section presents and analyzes the experimental results of the proposed method on the Kempsey flood case study. 
The content includes: a description of 2D and 3D evaluation metrics, comparative analysis of 2D flood segmentation performance, a detailed assessment of the HOTA strategy, and the effectiveness of the 3D flood depth modeling approach. 
Through these experiments and analyses, the proposed method is thoroughly validated in terms of its effectiveness and advantages in practical flood segmentation and depth reconstruction tasks.
\subsection{Evaluation Metrics}
\label{subsec:metrics}
To comprehensively evaluate the performance of SegFormer, HOTA, and the Flood Depth-Boundary Joint Optimization in 2D flood segmentation and 3D flood depth estimation, the following metrics are adopted:

For 2D segmentation, this work uses the following metrics: Accuracy (Acc), Precision, Recall, F1 Score (F1), and Intersection over Union (IoU).



For the 3D flood map, since ground-truth flood depth data are rarely available for real events~\cite{Poterek2025INFLOS}, the final 3D flood map is evaluated using boundary error metrics. 
The boundary error is defined as the mean absolute boundary error ($\mu_{\text{bnd}}$), as follows (Eq.~\ref{eq:metrics_3d}):
\begin{equation}
    \begin{aligned}
    \mu_{\text{bnd}} &= \frac{1}{N} \sum_{i=1}^{N} |d_i| \\
    \end{aligned}
    \label{eq:metrics_3d}
\end{equation}
where $d_i$ is the error between the predicted depth and reference depth (0 m) at the $i$-th flood boundary point, and $N$ is the total number of boundary points.

\subsection{2D Flood Mapping Results}
Building on the evaluation metrics outlined in Sec.~\ref{subsec:metrics}, this section compares the performance of SegFormer and U-Net on the STURM dataset (S2) and the Kempsey flood event dataset. 
In addition, we compare the performance of both models on the Kempsey dataset under varying HOTA configurations. 
This analysis provides further insight into the effectiveness of the HOTA strategy.

\textbf{SegFormer Vs U-net.}
As shown in Table~\ref{tab:models_sturm}, SegFormer consistently outperforms U-Net across all evaluation metrics. 
On the STURM S2 dataset, SegFormer achieves an IoU of 81.04\%, which is 5.35 percentage points higher than the U-Net baseline (75.69\%). 
\begin{table}[htbp]
    \centering
    \caption{SegFormer and U-net performance without HOTA.}
    \label{tab:models_sturm}
    \begin{tabular}{llccccc}
        \hline
        Dataset & Model & Accuracy & Precision & Recall & F1-Score & IoU \\
        \hline
        \multirow{3}{*}{\shortstack[c]{\textbf{STURM}\\(S2)}}
        & \textbf{U-net}~\cite{Notarangelo2025STURM} & --      & 72.59 & 82.49 & 77.22 & -- \\
        & \textbf{U-net}                             & 91.64  & 88.68 & 83.78 & 86.16&75.69 \\
        & \textbf{SegFormer}                         & \textbf{93.69} & \textbf{92.39} & \textbf{86.83} & \textbf{89.52} & \textbf{81.04} \\
        \hline
        \multirow{2}{*}{\textbf{Kempsey}}
        & \textbf{U-net}                             & 95.36  & 85.37 & 83.50 & 84.43 & 73.05 \\
        & \textbf{SegFormer}                         & \textbf{95.90} & \textbf{85.98} & \textbf{92.17} & \textbf{88.97} & \textbf{80.13} \\
        \hline
    \end{tabular}
\end{table}
This improvement can be attributed to the MiT architecture of SegFormer, which leverages self-attention mechanisms to capture long-range pixel dependencies. 
In contrast, the convolutional operations in U-Net are limited by a local receptive field. 
For flood segmentation tasks, the ability to model global context enables SegFormer to better capture the spatial continuity of large-scale flood regions, resulting in more accurate boundary delineation.

\textbf{HOTA.}
The different effects of the HOTA strategy on the two model architectures are illustrated in Tab.~\ref{tab:hota_comparison}. 
SegFormer demonstrates a consistent performance improvement under HOTA, with IoU gradually increasing from 80.13\% to 83.97\%. The gains are relatively uniform across all HOTA configurations. 
In contrast, U-Net achieves its greatest improvement when the HT is introduced, with IoU rising sharply from 73.05\% to 78.79\%, an overall increase of 5.74 percentage points. 
This indicates that CNN architectures rely more on explicit multi-scale feature fusion, whereas Transformer-based architectures can more effectively leverage the contextual information provided by hierarchical and overlapping tiles.
\begin{table}[htbp]
    \centering
    \begin{threeparttable}
    \caption{Models Performance under different HOTA configurations.}
    \label{tab:hota_comparison}
    \begin{tabular}{l|llccccc}
    \hline
    Dataset & Model & Configuration & Accuracy & Precision & Recall & F1-Score & IoU \\
    \hline
    \multirow{10}{*}{\shortstack{\textbf{Kempsey} \\ \textbf{Flood}}}
    & \multirow{5}{*}{\textbf{U-Net}}
    & No-HOTA & 95.36 & \textbf{85.37} & 83.50 & 84.43 & 73.05 \\
    & & OT & 95.73 & 84.94 & \textbf{93.19} & 86.02 & 75.47 \\
    & & HT        & 96.30 & 85.35 & 91.10 & 88.13 & 78.79 \\
    & & HOTA       & \textbf{96.41} & 84.56 & 92.77 & \textbf{88.67} & \textbf{79.64} \\
    \cline{2-8}
    & \multirow{5}{*}{\textbf{SegFormer}}
    & No-HOTA & 95.90 & 85.98 & 92.17 & 88.97 & 80.13 \\
    & & OT & 96.65 & 88.95 & 92.84 & 90.85 & 83.24 \\
    & & HT        & 97.11 & 88.63 & 92.74 & 90.64 & 82.88 \\
    & & HOTA       & \textbf{97.33} & \textbf{89.85} & \textbf{92.77} & \textbf{91.28} & \textbf{83.97} \\
    \hline
    \end{tabular}
    \begin{tablenotes}
        \footnotesize
        \item\textbf{Note:} OT = Overlap-Tiling; HT = Hierarchical Tiling.
    \end{tablenotes}
    \end{threeparttable}
\end{table}

\begin{figure}[htbp]
    \centering
    \includegraphics[width=0.8\textwidth]{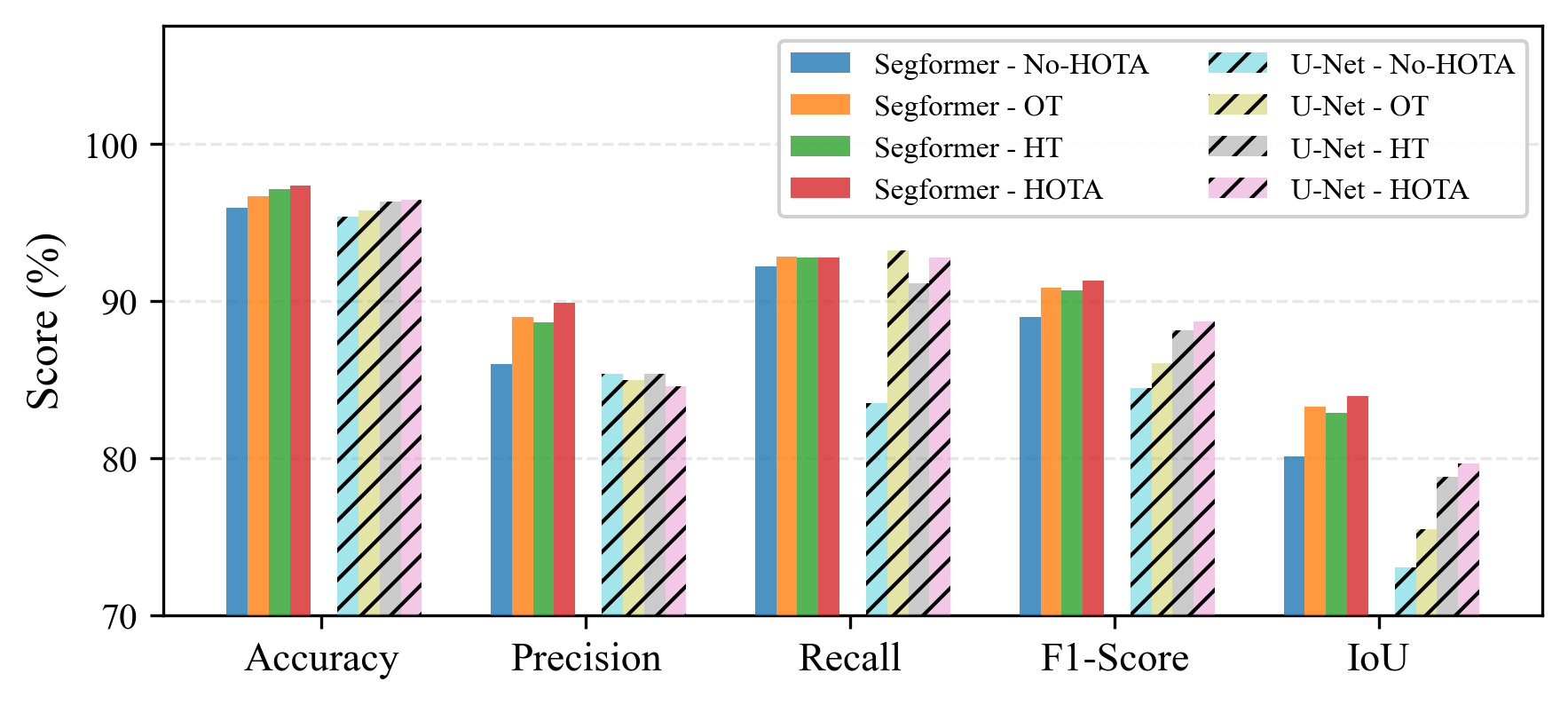}
    \caption{Comparison of metrics under different HOTA configurations.}
    \label{fig:metric_comparison}
\end{figure}

When visualizing these performance metrics, as shown in Fig.~\ref{fig:metric_comparison}, it is noteworthy that the two architectures exhibit different stability characteristics under the HOTA strategy.
SegFormer maintains a stable recall rate (92.17\%-92.77\%), while U-Net shows relatively stable precision but much larger fluctuations in recall (83.50\%-92.77\%). 

This reflects the ability of SegFormer's global attention mechanism to consistently capture flood features under varying input configurations, 
while U-Net's local feature extraction, although ensuring prediction consistency, is more sensitive to changes in context.

\begin{figure}[htbp]
    \centering
    \begin{subfigure}[b]{0.51\textwidth}
        \centering
        \includegraphics[width=\textwidth]{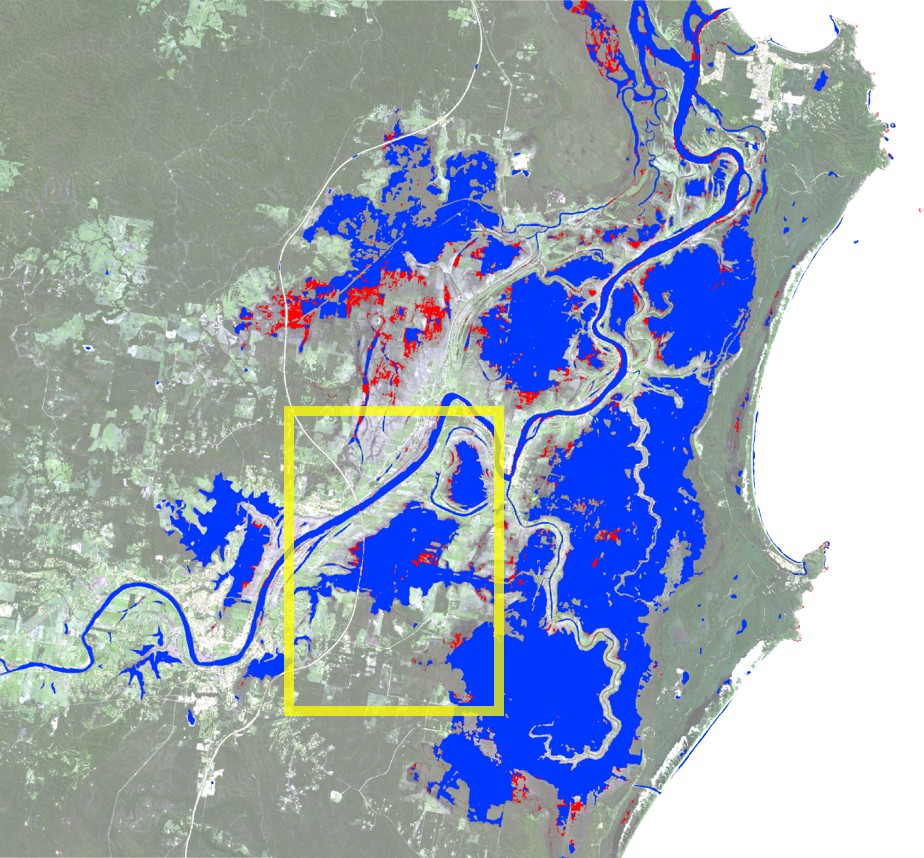}
        \caption{Full 2D map}
        \label{fig:2D_full}
    \end{subfigure}
    \begin{subfigure}[b]{0.32\textwidth}
        \centering
        \includegraphics[width=\textwidth]{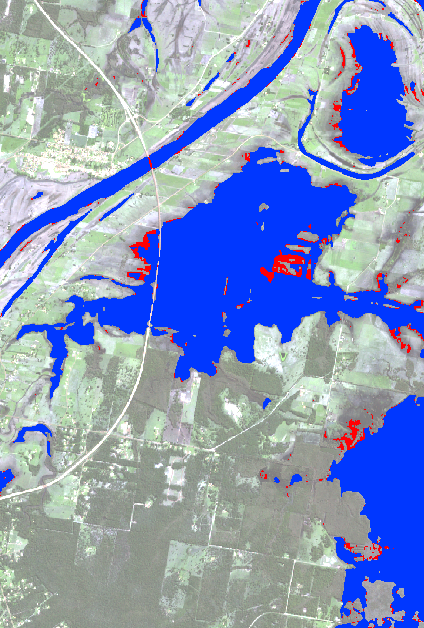}
        \caption{2D details}
        \label{fig:2D_detail}
    \end{subfigure}
    \caption{2D flood mapping results.}
    \label{fig:2D}
\end{figure}

Fig.~\ref{fig:2D} presents the 2D segmentation results of the SegFormer model with full HOTA configuration. 
The blue regions indicate the predicted water by SegFormer, while the red areas show missed detections. 
It can be seen that SegFormer with HOTA achieves good segmentation performance for both narrow river channels and large flood regions. 
However, some missed detections still occur along the boundaries of large water bodies and in fragmented flooded areas.

\begin{figure}[htbp]
    \centering
    \begin{subfigure}[b]{0.71\textwidth}
        \centering
        \includegraphics[width=\textwidth]{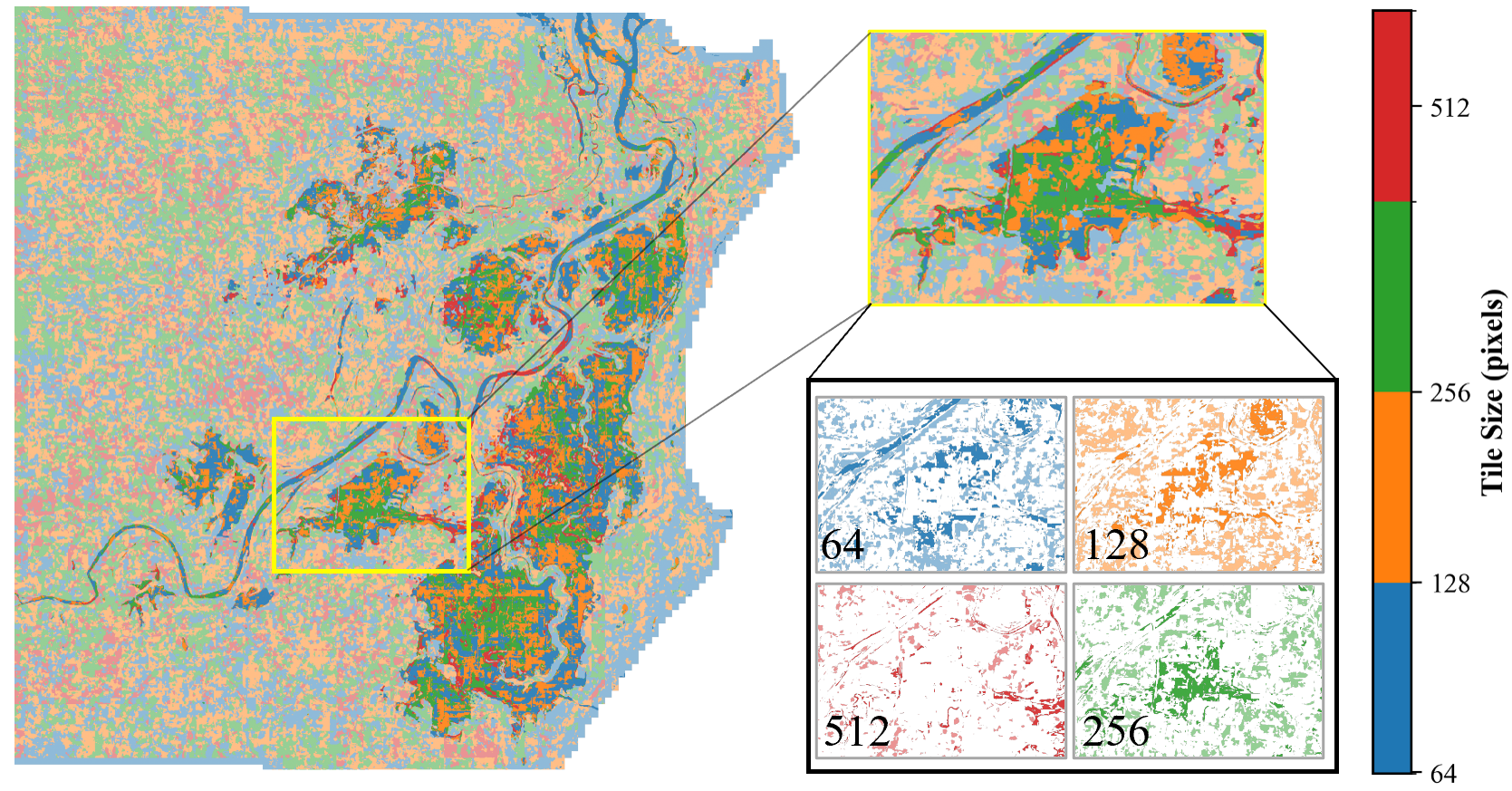}
        \caption{Scale contribution map}
        \label{fig:Scale_contribution}
    \end{subfigure}
    \vspace{0.5em} 
    \begin{subfigure}[b]{0.27\textwidth}
        \centering
        \includegraphics[width=\textwidth]{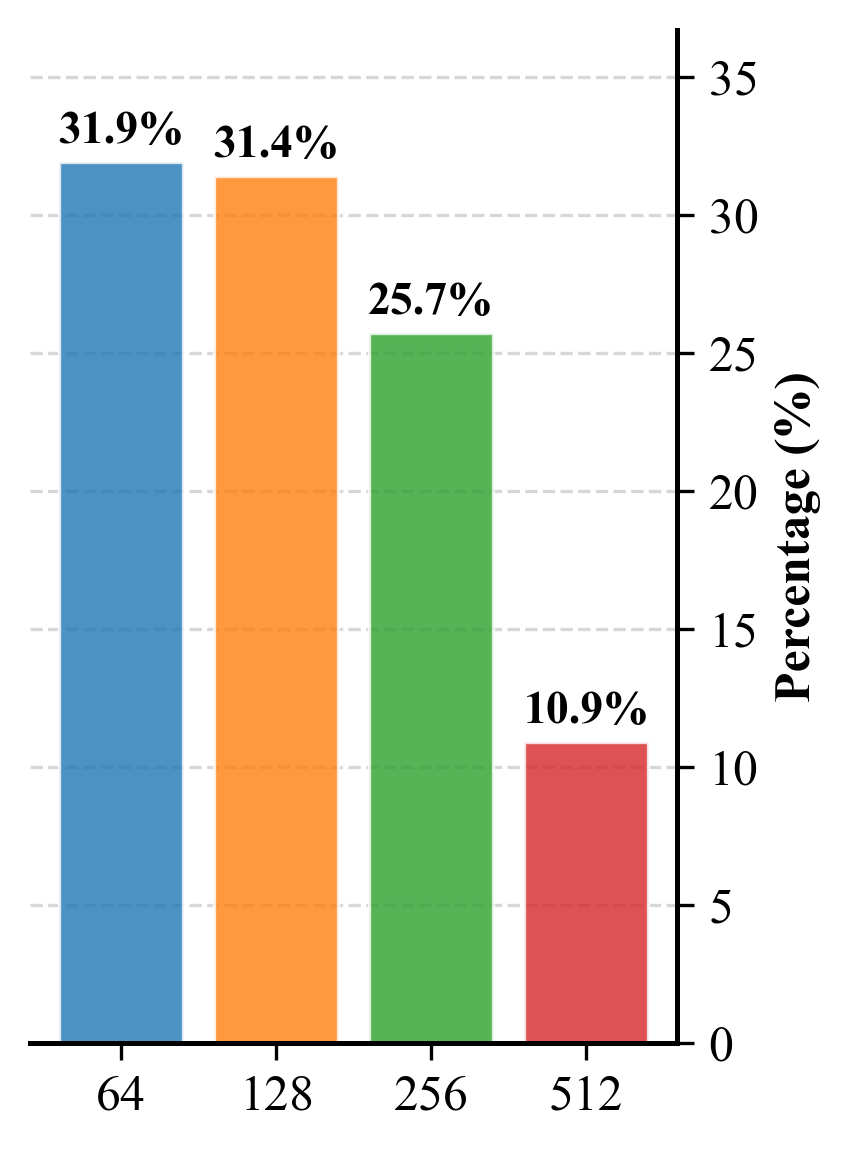}
        \caption{Distribution}
        \label{fig:Distribution}
    \end{subfigure}
    \caption{Comparison of scale-related visualizations.}
    \label{fig:HOTA_combine}
\end{figure}

The multi-scale contribution to the final 2D maps is shown in Fig.~\ref{fig:HOTA_combine}.
Blue regions represent contributions from the 64-pixel scale, orange indicates 128-pixel, green for 256-pixel, and red for 512-pixel tiles.

As shown in Fig.~\ref{fig:Distribution}, the 64-pixel and 128-pixel small-scale tiles contribute 63\% of the final prediction, while the 512-pixel tiles account for only 11\%. 
The low contribution of the largest scale may be related to the resolution mismatch with the 128-pixel tiles used during model training. 
The local zoomed-in view in Fig.~\ref{fig:Scale_contribution} shows that the 64-pixel scale primarily contributes to fine river channels and flood boundaries, while the 128-pixel and 256-pixel scales are most effective in the central, larger flooded regions. 
This distribution aligns with the nature of flood boundaries, which require fine analysis.

In summary, small-scale tiles are mainly responsible for capturing boundary details, whereas large-scale tiles provide global contextual constraints. 
Together, they ensure the spatial continuity of the segmentation results.

\subsection{3D Flood Mapping Results}
\begin{figure}[htbp]
    \centering
    \begin{subfigure}[b]{0.45\textwidth}
        \centering
        \includegraphics[width=\textwidth]{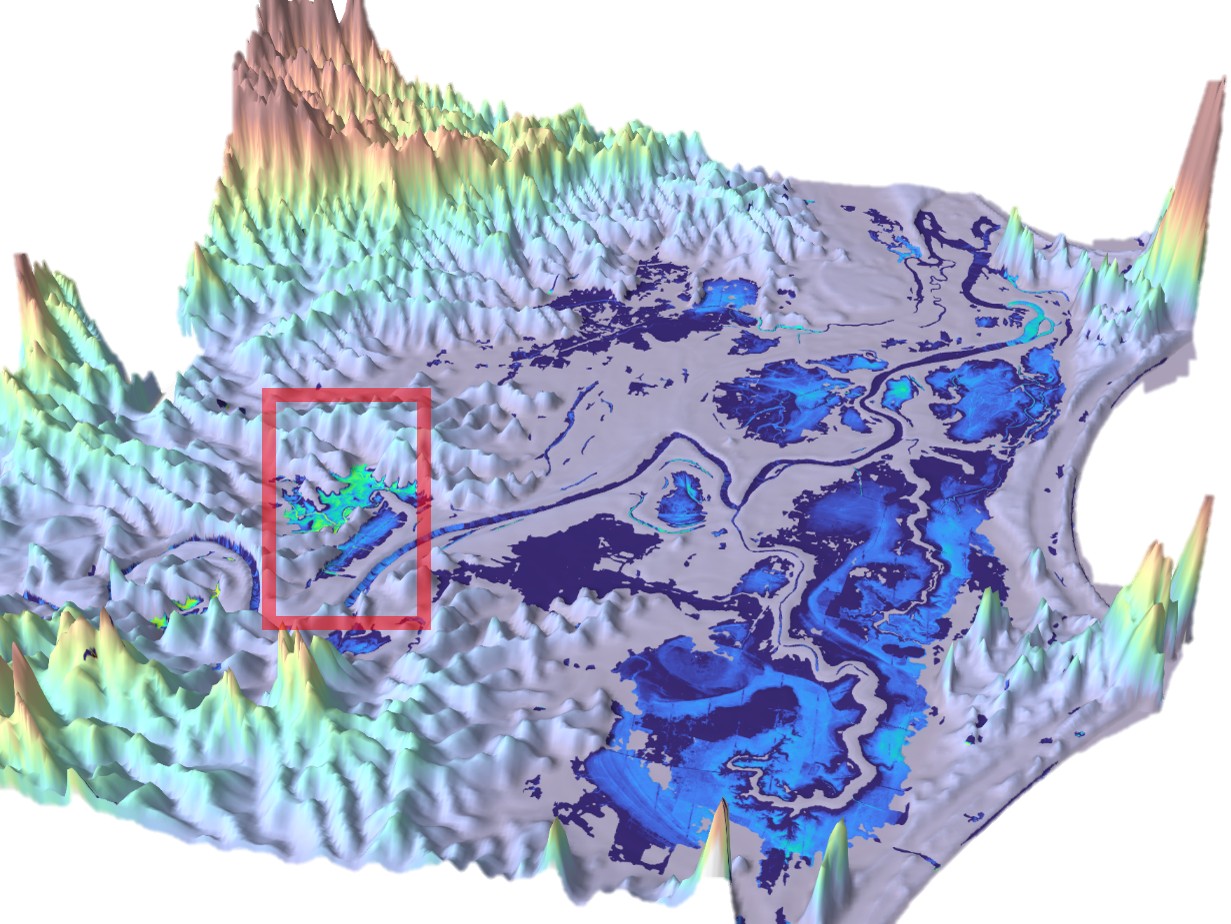}
        \caption{Full 3D view}
        \label{fig:3d_result_full}
    \end{subfigure}%
    \hspace{0.2em} 
    \begin{subfigure}[b]{0.18\textwidth}
        \centering
        \includegraphics[width=\textwidth]{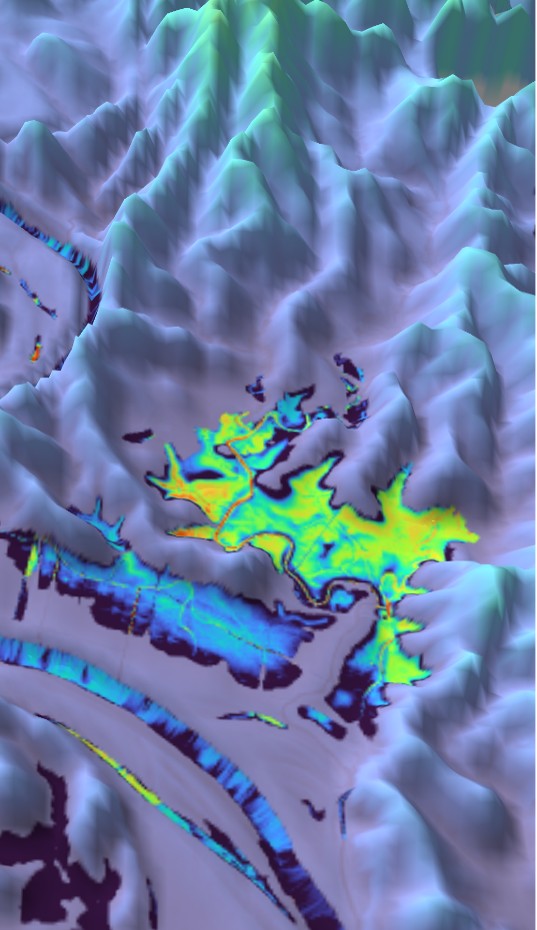}
        \caption{3D details}
        \label{fig:3d_result_detail}
    \end{subfigure}
    \begin{subfigure}[b]{0.75\textwidth}
        \centering
        \includegraphics[width=0.95\textwidth]{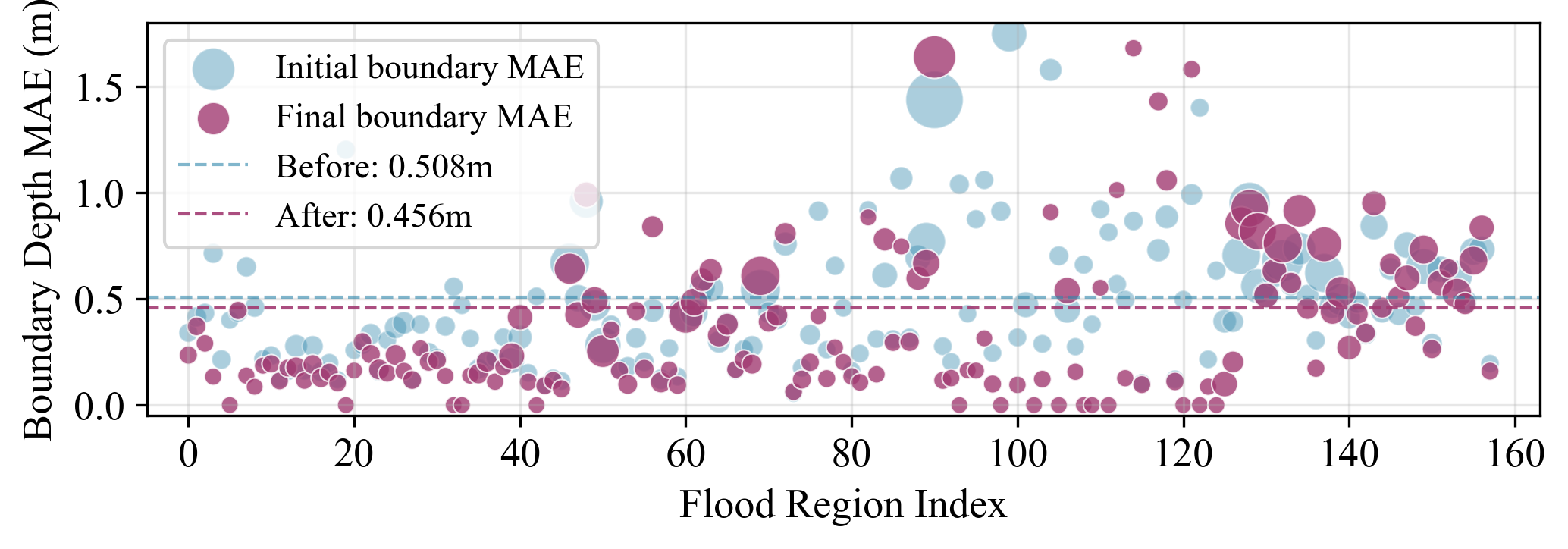}
        \caption{Comparison of boundary MAE before and after optimization.}
        \label{fig:3d_result_bubble}
    \end{subfigure}
    \caption{3D flood mapping results.}
    \label{fig:3d_result}
\end{figure}

The estimated 3D flood depth results are visualized in Fig.~\ref{fig:3d_result}. 
As shown in Fig.~\ref{fig:3d_result_detail}, it shows a zoomed-in view of the region highlighted by the red box in Fig.~\ref{fig:3d_result_full}. 
Fig.~\ref{fig:3d_result_bubble} compares the mean absolute error (MAE) of the boundaries depth before and after optimization: blue bubbles show the initial results, while purple bubbles indicate the optimized ones.
The size of each bubble corresponds to the number of boundary pixels in the flood region.

As shown in Fig.~\ref{fig:3d_result_detail}, the estimated 3D flood is concentrated in low-lying gullies and depressions within complex terrain, which is generally consistent with physical expectations. 
The comparison in Fig.~\ref{fig:3d_result_bubble} demonstrates that the boundary MAE improved from 0.508m to 0.456m after optimization, a reduction of approximately 10\%. 
Considering the vertical accuracy of the DEM data is 0.3~m, the optimized result is reasonably accurate.
It is also noteworthy that the proposed method yields better improvements for smaller flood regions (with fewer boundary pixels), 
while the optimization effect is less pronounced for larger flood areas (with more boundary pixels). 
This may be because the terrain around the boundaries of larger flood regions is more complex, making it more challenging to identify a physically plausible flood surface elevation.

\section{Discussion}
This study demonstrates that the HOTA approach offers clear advantages for flood segmentation and 3D modeling. 
With its global attention mechanism, SegFormer consistently extracts relevant flood features across varying scales and complex scenes, outperforming traditional CNN-based models like U-Net. 
In the Kempsey case, SegFormer with HOTA improved 2D segmentation IoU by 3.9 percentage points (from 80\% to 84\%), surpassing the U-Net baseline (73\%) and highlighting its reliability in real-world scenarios.

The HOTA strategy enhances SegFormer's performance by enabling effective multi-scale context integration without altering the model or retraining. 
HOTA produces steady accuracy gains and is highly portable for other remote sensing tasks. Future work may focus on optimizing fusion strategies and validating HOTA with more complex targets.
For 3D flood modeling, DEM-based depth estimation with dual physical constraints ensures realistic water distribution. 
Experiments show the mean absolute boundary error can be reduced to under 0.5 meters, providing dependable spatial data for disaster response.

Some limitations remain. Current 2D segmentation relies on single-date imagery and cannot reflect flood dynamics. 
The 3D modelling method also depends on DEM quality, and boundary optimization may still struggle in complex terrains. 
Future research should incorporate multi-source and multi-temporal data, develop more robust boundary algorithms, and include uncertainty analysis to further strengthen the model's application in flood risk assessment.

\section{Conclusion}
This work proposes \textbf{HOTA} and a HOTA-based pipeline for 3D flood mapping. The presented pipeline integrates the HOTA strategy with the SegFormer model and a dual-constraint DEM-based depth estimation method, enabling high-precision, large-scale flood segmentation and water depth modeling. 
The HOTA strategy enhances segmentation performance on large remote sensing scenes, while the boundary and volume constraints ensure the physical plausibility of 3D flood depth estimates.

Experiments on the March 2021 flood in Kempsey, Australia, demonstrate that  HOTA combined with SegFormer significantly outperforms U-Net, improving the IoU from 73.05\% to 83.97\%. 
In addition, the pipeline achieves high-quality 3D flood reconstruction, highlighting its strong potential for real-world flood emergency management, disaster warning, and decision support.

Nevertheless, certain limitations remain.
The 3D method relies on high-quality DEM, and the adaptability of depth optimization in large-scale, complex terrain remains a challenge. 
Furthermore, a systematic analysis of model uncertainty is currently lacking, which affects its applicability for risk assessment.

Future research will focus on incorporating multi-source data (e.g., SAR, LiDAR), multi-temporal imagery, and developing more robust boundary handling algorithms to improve performance and generalizability in complex scenarios.
Additionally, uncertainty quantification methods will be introduced to enhance the reliability and practicality of model predictions, supporting broader adoption of 3D flood mapping in risk assessment and emergency response.

\subsubsection{Acknowledgements} Please place your acknowledgments at
the end of the paper, preceded by an unnumbered run-in heading (i.e.
3rd-level heading).

%
%
%
\bibliographystyle{splncs04}
\bibliography{ACPR_references}
%




\end{document}